# Hardware Acceleration of Lane Detection Algorithm: A GPU Versus FPGA Comparison

Mohamed Alshemi[1], Sherif Saif[2] and Mohamed Taher[3]

[1]Computer and Systems Engineering Department,
Ain Shams University, Cairo, Egypt
[1]STMicroelectronics, Cairo, Egypt
[2]Computer and Systems Department,
Electronics Research Institute, Cairo, Egypt
[3]Computer and Systems Engineering Department,
Ain Shams University, Cairo, Egypt

*Abstract*

*A Complete Computer vision system can be divided into two main categories: detection and classification. The Lane detection algorithm is a part of the computer vision detection category and has been applied in autonomous driving and smart vehicle systems. The lane detection system is responsible for lane marking in a complex road environment. At the same time, lane detection plays a crucial role in the warning system for a car when departs the lane. The implemented lane detection algorithm is mainly divided into two steps: edge detection and line detection. In this paper, we will compare the state-of-the-art implementation performance obtained with both FPGA and GPU to evaluate the trade-off for latency, power consumption, and utilization. Our comparison emphasises the advantages and disadvantages of the two systems.*

*Keywords*

*Lane Detection, Computer Vision, FPGA, GPU, CUDA.*

## 1. Introduction

The problem of lane detection has long been an important part of driver assistance systems (ADAS). As the first step in road segmentation, lane detection facilitates scene understanding for intelligent vehicles. For example, many vehicle functions require lane detection, such as lane departure warning (LDW), adaptive cruise control, lane centring, and lane change assist. There have been some advances in the lane detection problem in recent years.

Due to the computational complexity of the lane detection algorithm, power consumption can be high to achieve real-time performance in a lane detection solution. While power is not a concern for all designs, low power solutions are often required for embedded or edge computing solutions. In the past, computationally complex image processing has been performed using graphical processing units (GPUs), but due to advances in field programmable gate array (FPGA) technology these tasks have become more viable with lower power consumption FPGA devices.
So through this paper, two lane detection hardware implementation solutions are developed and compared. One solution is implemented on a ZYNQ-7 ZC706 Evaluation Board FPGA [1] and the other solution is implemented on Google Colab NVidia K80 Super GPU [2]. The implementation for both FPGA and GPU will be shown in Figure 1.

The implementation includes the following steps.

Step 1: Sobel edge detection for a gray image.
Step 2: Applying the image binarization.
Step 3: Obtaining the Hough matrix as a result of the Hough transform.
Step 4: MATLAB functions are used to process the Hough matrix to draw the lane lines.

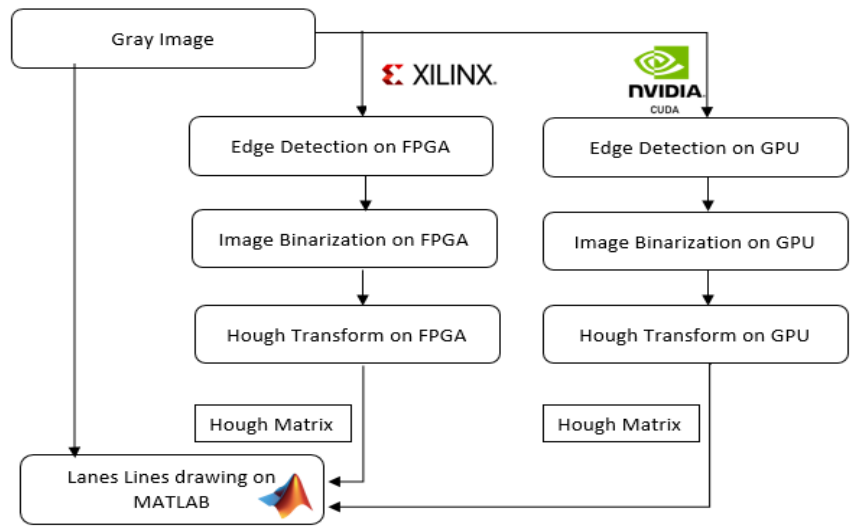

Figure 1.  Lane Detection implementation overview

## 2. THE ALGORITHM EMPLOYED FOR LANE DETECTION

Lane detection is divided into three parts [3].

### 2.1. Sobel Edge Detection

The Sobel edge detection algorithm is used to extract features from images like edges [4]. It works by calculating the image intensity gradient for each pixel within the image. It finds the direction of the largest change from light to dark and the change rating in that direction, so the result shows how the image changes abruptly or smoothly at each pixel, and therefore how much that pixel looks like an edge and it also shows how the orientation of that edge is. The output from applying a Sobel filter to a pixel on an edge is a vector that changes across the edge from darker to brighter values.

The Sobel filter uses two 3 x 3 matrix kernels as shown in Figure 2. One for calculating the changes in the horizontal direction, and the other for calculating the changes in the vertical direction. The two kernels are convolved with the original gray pixel image to calculate the derivative approximations. Where the approximated magnitude is

$|G| = |Sx * Img| + |Sy * Img|$  (1)

Figure 2. Sobel kernels coefficients for edge detection

## 2.2. Image Binarization

The purpose of image binarization is to binarize images before further processing. Following that, all of the image's pixels will either be 255 or 0 (a pixel is normally represented by 8 unsigned bits, making 255 the highest value in the 8-bit data format).

## 2.3. Hough Transform

After image binarization, the Hough transform is used to detect the two lines of the lane [4]. The Hough transform is the most important technique to find shapes in images. It has been specifically used to extract circles, lines, and random shapes. In other words, the Hough transform algorithm has been recognized as a promising technique for shape and motion analysis in images, so the Hough transform is an ideal algorithm for line detection [5]. A line is represented by the following equation:

$$y = a \cdot x + b \qquad (2)$$

In [r, θ] coordinates, a line can be:

$$r = x \cdot cos(\theta) + y \cdot sin(\theta) \qquad (3)$$

Where parameter θ is the angle of the line, and parameter r is the distance from the line to the origin. Thus, the line can be represented by a single point of Polar (r, θ) coordinate in Hough space. In this way, a line in the Cartesian (x, y) coordinate system can be mapped to a single point in (r, θ) Hough space, as indicated in Figure 3.

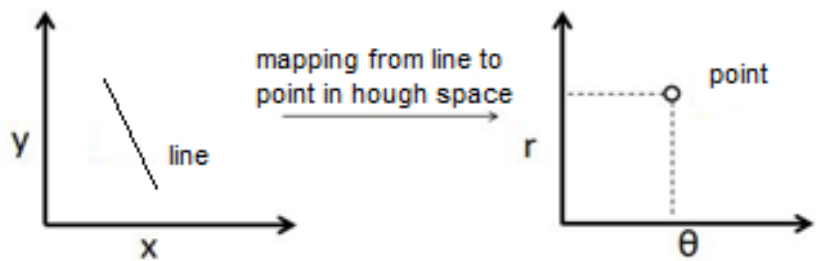

Figure 3. Mapping of single line to the Hough space

Conversely, a point in the (x, y) coordinate system represents all lines passing through that point. This way a point maps all possible lines that pass through it in Hough space. As shown in Figure 4, a point maps to a line that looks like a sinewave in the Hough space [5].

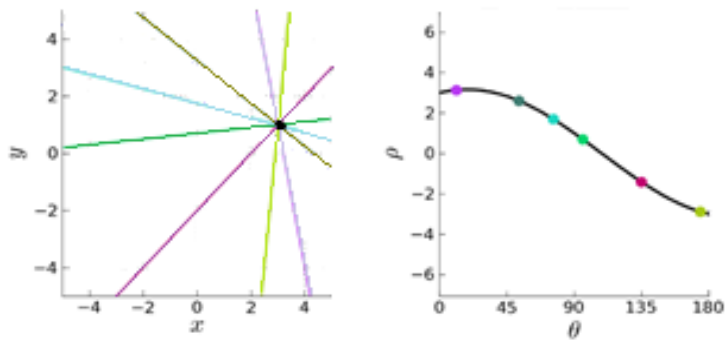

Figure 4. Mapping of crossed lines to the Hough space

To recognize straight lines inside an image, each point in the binarized generated image votes to its corresponding possible lines, which are pairs of (r, θ) in Hough space. After processing each pixel in the input image, an accumulator array is generated. The line with the largest count value denotes the line with the most pixels on it. The process of picking the line with the most pixel value is called Hough Peak which is done in Matlab.

## 3. HARDWARE IMPLEMENTATION ON FPGA

### 3.1. Sobel Edge Detection FPGA Implementation

After receiving the gray image, the first step is doing the zero padding for the original image to ensure that the output image size is the same as the input image.

A general architecture to perform kernel-based convolution computation is shown in Figure 5. The Sobel filter is implemented using adders and left shifters as Sobel filter parameters merely have three forms: 1, 2, and 0.

Replacing multipliers with left shifting by 1 to realize multiplication by 2 helps in reducing resource utilization on hardware (Area optimization).

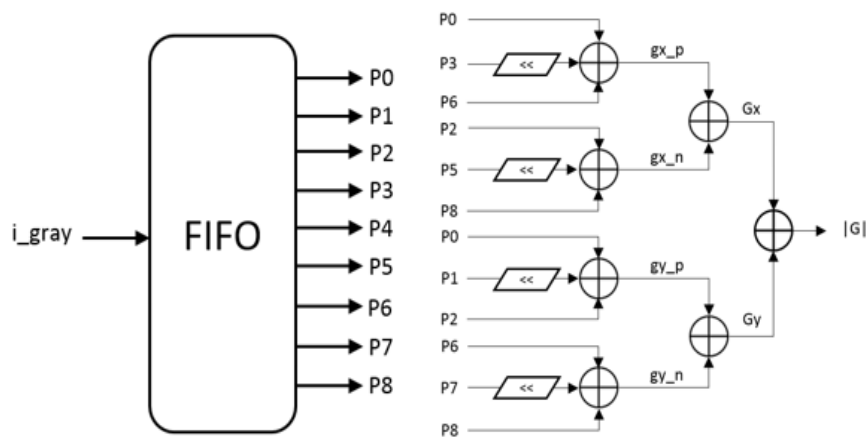

Figure 5. FPGA implementation of Sobel edge detection

### 3.2. Image Binarization FPGA Implementation

Image binarization on hardware is implemented by using a multiplexer and an unsigned comparator as shown in Figure 6.

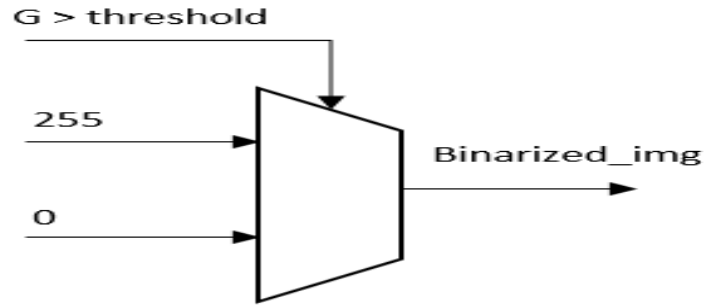

Figure 6. FPGA implementation of image binarization

### 3.3. Hough Transform FPGA Implementation

For each pixel equal to 255 after the binarization, lines from all possible directions are counted. The range of the line angles is [0, 180] and the distance from the origin for each angle is determined using the structure shown in Figure 7, which can be considered as the Hough transform's fundamental processing element. The degree resolution is 1 degree between angles, hence 180 such processing elements are needed to compute for 180 degrees range. The result of each processing element is accumulated and stored in memory [6].

Figure 7 shows that each processing element cell needs two multipliers, and a total of 180 such processing elements are used, resulting in using 360 multipliers. We further optimize the usage of multipliers by adjusting the properties of sine and cosine functions [7]. Such properties are expressed in the following equations:

$$\sin \theta = \sin(180 - \theta) \qquad (4)$$

$$\cos \theta = -\cos(180 - \theta) \qquad (5)$$

Where $\theta_1 + \theta_2 = 180$, $\sin(\theta_1) = \sin(\theta_2)$, and $\cos(\theta_1) = -\cos(\theta_2)$. In other words, once sin ($\theta_1$) and cos ($\theta_1$) are calculated, sin ($\theta_2$) and cos ($\theta_2$) can be done by simple add/minus operations. In this way, only 180 multipliers are used instead of 360, saving 50% of the total number of multipliers.

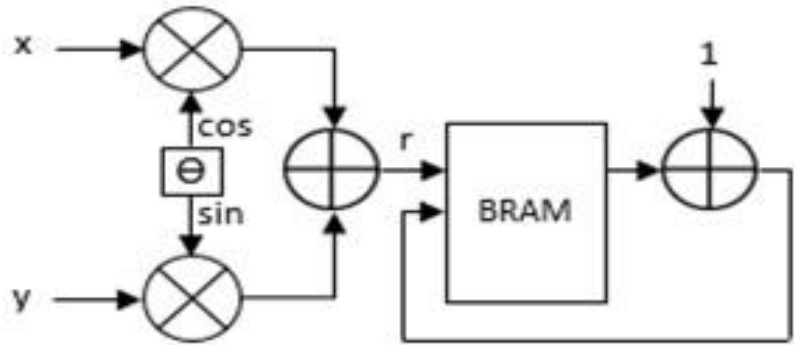

Figure 7. FPGA implementation of Hough transform

## 4. Hardware Implementation on GPU

Our GPU implementation was based on Google Colab NVidia Tesla K80 Super GPU connected to a host CPU via PCIe. The GPU has 4992 NVIDIA CUDA cores with a base clock of 562 MHz it has 12GB global memory operating on a memory clock of 1253 MHz CUDA was used to program the GPU.

With less modifications to C, NVidia's Compute Unified Device Framework (CUDA) offers a general architecture for GPUs that enables flexible programming. CUDA's view of a GPU has plentiful Streaming Multiprocessors. Each multiprocessor has a shared memory which is a fast local memory shared by its streaming processors. All multiprocessors also have access to a global or device memory. A thread block can be formed by many threads on a streaming multiprocessor. A thread block maps to a Streaming Multiprocessor physically although the threads number inside a block can be bigger than the number of Streaming processors, in which case threads are split into multiple warps where only one warp is active at a time. Active threads can execute the same instruction in parallel.

The components of the Lane Detection must be carefully mapped to the GPU computing architecture in order to make optimum use of GPU computing resources and to reduce the access cost of device memory access.

Two principles guided our mapping:

1. Avoid unnecessary device memory access and prefer shared memory for repeated memory access and
2. Avoid branches to keep all the processors doing useful computation work.

Currently available GPUs have a limited amount of shared memory per multiprocessor, with the majority of them providing roughly 48KB of shared memory per block [8].

### 4.1. Sobel Edge Detection GPU Implementation

After receiving the gray image, we start with doing the zero padding for the original image and store it inside the device memory then we use a 2D grid which contains a 2D Block with a size of 16*16 threads, as in the following equation

$$\text{grid} = \text{dim3}((\tfrac{\text{width}+\text{BLOCK}_{SIZE}-1}{\text{BLOCK}_{SIZE}}), (\tfrac{\text{height}+\text{BLOCK}_{SIZE}-1}{\text{BLOCK}_{SIZE}}))  \qquad (6)$$

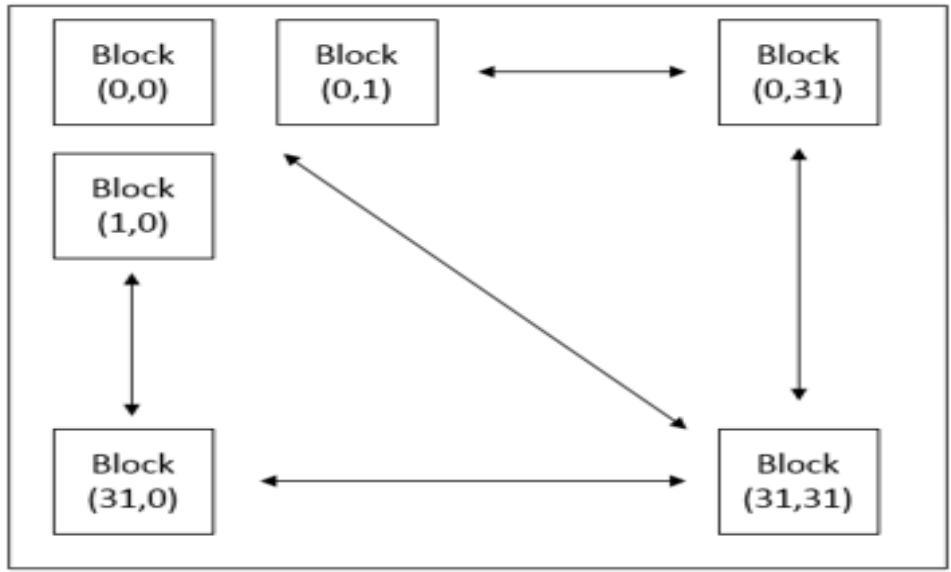

Figure 8. GPU block diagram of grid thread blocks

Where *width* and *height* are the dimensions of the original gray image. Each block will perform the Sobel operator convolution on the gray image and all the grid blocks will perform the computations in parallel.

### 4.2. Image Binarization GPU Implementation

Each block thread will be responsible for the threshold of the post-Sobel pixels to be 255 or 0.

### 4.3. Hough Transform GPU Implementation

The Hough transform voting space (r, θ) is divided into partitions for each θ and the voting method for each partition is performed concurrently. A thread block in the GPU implementation [9], is assigned to each θ where threads in the thread block concurrently vote for input post-binarization pixels. Partitioning the voting space for each θ where each voting space can be stored in the shared memory. Therefore, the voting mechanism is performed to the voting space in the block shared memory. In the GPU implementation, the voting space is partitioned into 180 spaces where thread blocks can vote in parallel. The values of the trigonometric functions cos θ and sin θ are initially computed in each thread block since the values are commonly used for every thread in a thread block. Then threads read the coordinates of input pixels stored in the global memory.

Some threads may vote to the same *r* simultaneously in the parallel voting procedure. To avoid it, we use the atomic add operation supported by CUDA [10].

## 5. EXPERIMENTAL RESULTS

The two systems' performance was evaluated based on latency, power consumption, and resource utilization.

1- Latency was measured for both implementations between the first valid pixel of the image entering the system and the generation of the Hough Matrix.
2- Power was measured for both implementations as an average power using the devices' specific tools.
3- Memory resource utilization for both implementations was measured for the Hough transform as BRAM memory in FPGA and shared memory in GPU.

The results obtained for the proposed lane detection algorithm for the gray image in Figure 9 and the Sobel detection and binarization output for both implementations as in Figure 10. Some MATLAB built-in functions were used to detect the *Hough peaks* as in Figure 11. For the Hough Matrix generated from both FPGA and GPU and *Hough lines* as in Figure 12 to draw the lane lines.

Table 1 shows the resource utilization of the Lane Detection Algorithm on ZYNQ-7 ZC706 Evaluation Board FPGA for 100 MHz frequency whereas Table 2 shows the performance comparison of the Lane Detection Algorithm between FPGA and GPU.

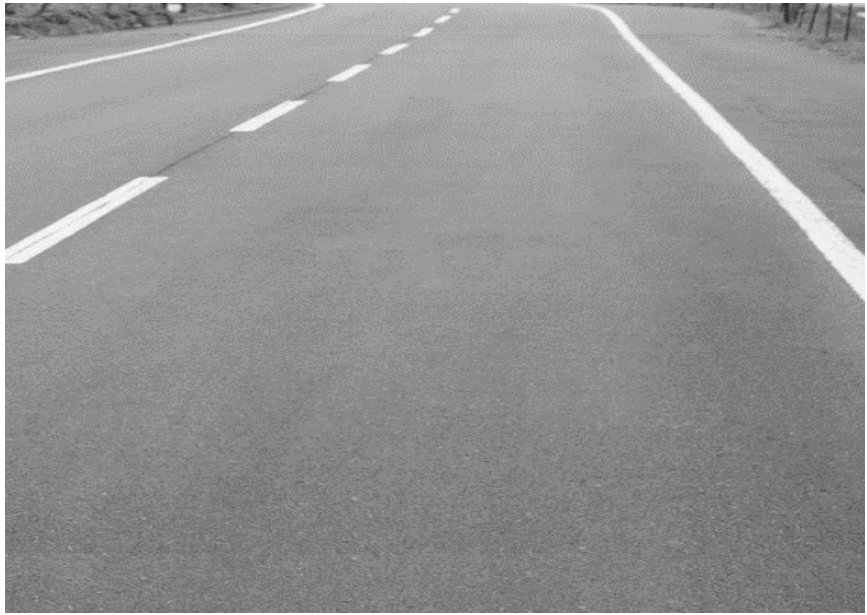

Figure 9. [512 x 512] pixel gray image

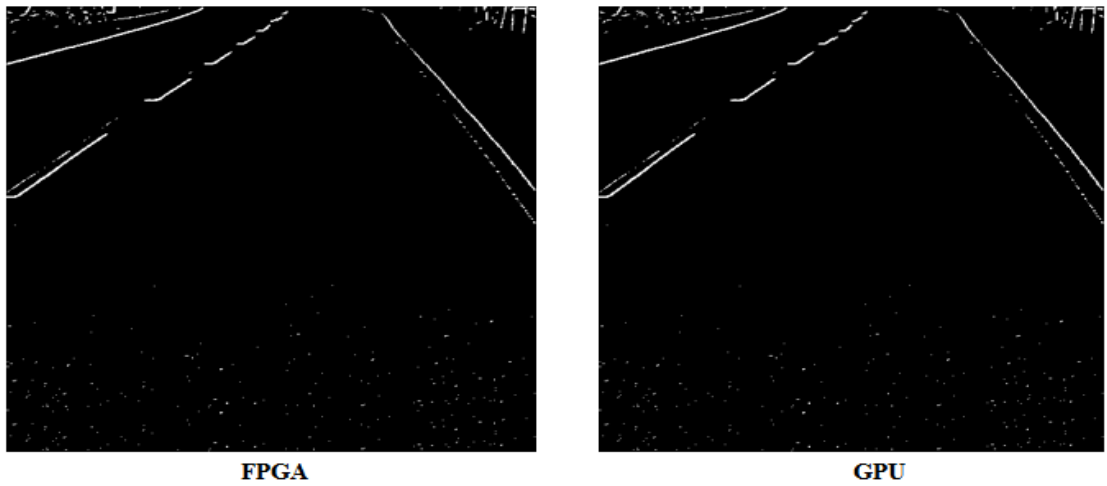

Figure 10. Image after Sobel detection and binarization

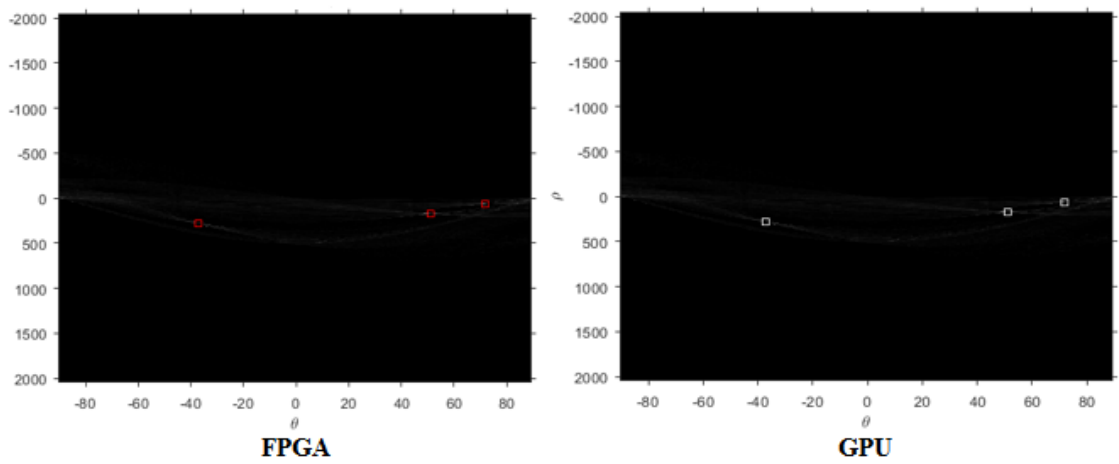

Figure 11. Matlab Hough peaks for Hough Matrix

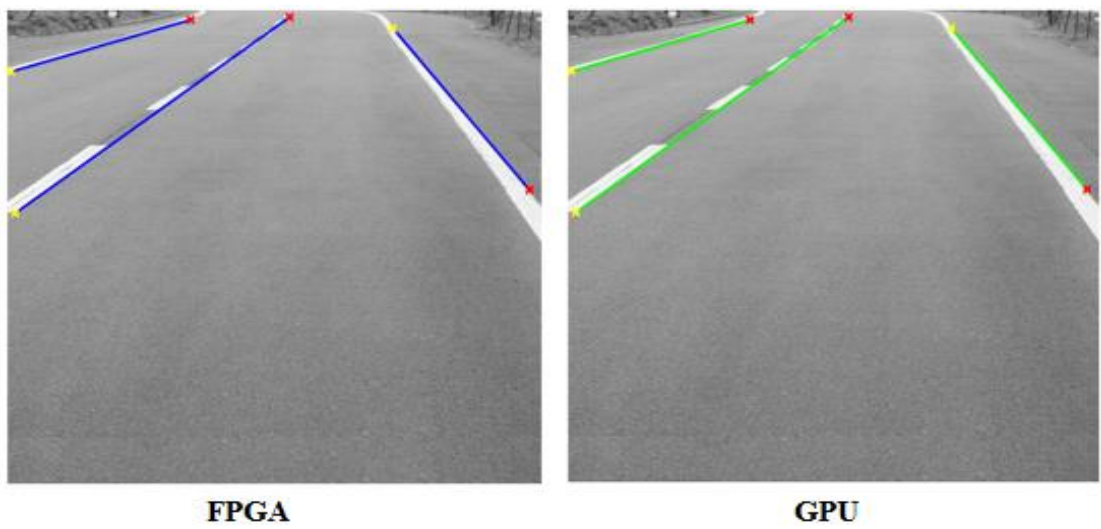

Figure 12. Matlab drawing of lane lines

Table 1. The resources utilization on ZYNQ-7 ZC706 FPGA

| Resources | Utilization | Available | Percentage % |
|---|---|---|---|
| **LUT** | 20022 | 218600 | 9.159195 |
| **FF** | 9439 | 437200 | 2.158966 |
| **BRAM** | 180 | 725 | 24.8 |
| **DSP** | 180 | 900 | 20.0 |

Table 2. the performance comparison of THE Lane Detection Algorithm between FPGA and GPU

|  | GPU | FPGA |
|---|---|---|
| **Model** | *Tesla K80* | *ZYNQ-7 ZC706 Evaluation Board* |
| **latency** | *4.874000ms* | *2.62ms* |
| **Power Consumption** | *74.22 W* | *1.619 W* |
| **utilization** | *6.5 MB memory* | *2.88 MB memory* |

## 6. CONCLUSION

In this paper, two implementations of lane detection were developed on two different forms of hardware acceleration. The two forms of hardware acceleration used were an FPGA and a GPU and tests were performed using only the acceleration portion of the lane detection algorithm.

Tests show low latency for FPGAs, low relative memory resource consumption for GPUs, and low power consumption for FPGAs. GPUs handle resolution scaling as well or better than FPGAs in terms of memory resource consumption. FPGAs handle resolution scaling as well or better than GPUs in terms of latency. If your project will be implemented in this work and you need high resolution which may be needed later in the project then GPU is probably the better choice. If performance and latency are critical and resolution may be limited, FPGAs are probably a better choice.

Finally, the results showed that FPGA has better results than the GPU for [512x512] pixel images for Latency, power consumption, and resource utilization but GPU will handle the resolution scaling better than the FPGA with regards memory resource consumption.

## 7. FUTURE WORK

In future work, it would be beneficial to use canny edge detector instead of Sobel edge detector and analyze its effect on the performance of the output image after the edge detection.

## REFERENCES

[1] Xilinx Inc, "ZC706 Evaluation Board for the Zynq-7000 XC7Z045 SoC User Guide (v1.8)." 2019.
[2] S. Mittal and J. S. Vetter, "A survey of CPU-GPU heterogeneous computing techniques," ACM Computing Surveys, vol. 47, no. 4, pp. 1–35, 2015.
[3] C. Lee and J.-H. Moon, "Robust lane detection and tracking for real-time applications," IEEE Transactions on Intelligent Transportation Systems, vol. 19, no. 12, pp. 4043–4048, 2018.
[4] R. C. Gonzalez and R. E. Woods, Digital Image Processing. Uttar Pradesh, India: Pearson, 2018.
[5] Y. Zhou, "Computer Vision System-On-Chip Designs for Intelligent Vehicles," Ph.D. dissertation, UC Davis, 2018.
[6] D. Northcote, L. H. Crockett, and P. Murray, "FPGA implementation of a memory-efficient Hough parameter space for the detection of lines," 2018 IEEE International Symposium on Circuits and Systems (ISCAS), 2018.

[7]   X. Zhou, Y. Ito, and K. Nakano, "An efficient implementation of the Hough transform using DSP slices and block rams on the FPGA," 2013 IEEE 7th International Symposium on Embedded Multicore Socs, 2013.

[8]   "Cuda C++ Programming Guide," NVIDIA Documentation Center. [Online]. Available: https://docs.nvidia.com/cuda/cuda-c-programming-guide/index.html. [Accessed: 04-Nov-2022].

[9]   S. Bagchi and T.-J. Chin, "Event-based star tracking via multiresolution progressive Hough transforms," 2020 IEEE Winter Conference on Applications of Computer Vision (WACV), 2020.

[10]  X. Zhou, N. Tomagou, Y. Ito, and K. Nakano, "Implementations of the hough transform on the embedded multicore processors," International Journal of Networking and Computing, vol. 4, no. 1, pp. 174–188, 2014.